\documentclass[runningheads]{llncs}
\usepackage{graphicx}
\usepackage{amsmath,amssymb}
\usepackage{pifont}
\usepackage{algorithmic}
\usepackage{graphicx}
\usepackage{textcomp}
\usepackage{xcolor}
\usepackage{multirow}
\usepackage{tabularx,ragged2e}
\usepackage{tabulary}
\usepackage{adjustbox}
\usepackage{arydshln}
\usepackage{url}
\usepackage{hhline}
\usepackage{multicol}
\usepackage{cite}
\usepackage{booktabs}
\usepackage{mathtools}

\usepackage{cleveref}
\usepackage{footmisc}
\usepackage{graphicx}
\usepackage{arydshln}
\usepackage{comment}
\usepackage[flushleft]{threeparttable}

\newcommand{\cmark}{\ding{51}}%
\begin{document}
\title{Exploiting Transformer-based Multitask Learning for the Detection of Media Bias in News Articles}
\titlerunning{Transformer-based Multitask Learning on Media Bias}
%
\author{Timo Spinde\inst{1}\orcidID{0000-0003-3471-4127} \and
Jan-David Krieger\inst{2}\orcidID{0000-0002-5360-2078} \and
Terry Ruas\inst{1}\orcidID{0000-0002-9440-780X} \and
Jelena Mitrović\inst{3}\orcidID{0000-0003-3220-8749} \and
Franz G{\"o}tz-Hahn\inst{4}\orcidID{0000-0003-3465-5040 } \and
Akiko Aizawa\inst{5}\orcidID{0000-0001-6544-5076} \and 
Bela Gipp\inst{1}\orcidID{0000-0001-6522-3019}}
%
%
\institute{University of Wuppertal, Germany \\ \email{\{firstname.lastname\}@uni-wuppertal.de}
\and
University of Konstanz, Germany \\ \email{Jan-David.Krieger@uni-konstanz.de}
\and
University of Passau, Germany \\ Institute for Artificial Intelligence Research and Development of Serbia \\ \email{jelena.mitrovic@Uni-Passau.de}
\and
University of Kassel, Germany \\ \email{franz.goetz-hahn@uni-kassel.de}
\and
NII Tokyo, Japan \\ \email{aizawa@nii.ac.jp}}

\maketitle              
\begin{abstract}
 Media has a substantial impact on the public perception of events. A one-sided or polarizing perspective on any topic is usually described as media bias. One of the ways how bias in news articles can be introduced is by altering word choice. Biased word choices are not always obvious, nor do they exhibit high context-dependency. Hence, detecting bias is often difficult. We propose a Transformer-based deep learning architecture trained via Multi-Task Learning using six bias-related data sets to tackle the media bias detection problem. Our best-performing implementation achieves a macro $F_{1}$ of 0.776, a performance boost of 3\% compared to our baseline, outperforming existing methods. Our results indicate Multi-Task Learning as a promising alternative to improve existing baseline models in identifying slanted reporting.

\keywords{Media Bias \and Text Analysis \and Multi-Task Learning \and News Analysis}
\end{abstract}
\section{Introduction}\label{sec:introduction}
Media bias, i.e., slanted news coverage, has the potential to drastically change the public opinion on any topic \cite{SpindeBABE}. One of the forms bias can be expressed by is by word choice, e.g., depicting any content in a non-neutral way \cite{recasens2013a}. Detecting and highlighting media bias might be relevant for media analysis and mitigate the effects of biased reporting. To date, only a few research projects focus on the detection and aggregation of bias \cite{lim2020a,chen_analyzing_2020}. One of the reasons that make the creation of automated methods to detect media bias a complex task is often the subtle nature of media bias, which represents a challenge for quantitative identification methods \cite{fan2019a,lim2020a, Spinde2021Embeddings, spinde_how_2021}. While many current research projects focus on collecting linguistic features to describe media bias \cite{hube2018detecting, Spinde2021, Spinde2020INRA, recasens2013a}, we propose a Transformer-based \cite{vaswani2017attention} architecture for the classification of media bias. Similar models have recently shown to achieve performance increases in the media bias domain, e.g., sentence-level bias detection \cite{chen_analyzing_2020, hube2019neural, SpindeBABE, spinde2021g}. However, so far, they rely on very limited resources. Data sets with bias gold standard annotations are, to date, only scarcely available, and exhibit various weaknesses, such as low inter annotator agreement, small size, or no information about the annotator background \cite{SpindeBABE, spinde_towards_2021, spinde_tassy_2021}. Additionally, state-of-the-art neural language models usually require large amounts of training data to yield meaningful representations \cite{devlin2018bert, ruder2017overview}, which are incompatible with the size of current media bias data sets \cite{spinde2021mbic, fan2019a}. To mitigate the lack of suitable data sets, our model incorporates Multi-Task Learning (MTL) \cite{ruder2017overview}, which allows for increasing performance by sharing model representations between related tasks \cite{sun2019fine,huo2020utilizing, liu2019multitask}. The use of cross-domain data sets in our model is particularly relevant for the media bias domain as multiple sources can provide a more robust model. To the best of our knowledge, the MTL paradigm has not been explored in existing work on media bias. Our  research question is therefore to assess whether MTL can improve models to classify media bias automatically. 

The main contribution of this paper is to incorporate Transformer-based MTL into a system to identify sentence-level media bias automatically. We exploit MTL in the media bias context by computing multiple models based on different combinations of auxiliary training data sets (\cref{sec:related_work}). All our models, data, and code are publicly available on \url{https://bit.ly/3cmiQgB}.


\section{Related Work}\label{sec:related_work}
While there are multiple forms of media bias, e.g., bias by personal perception or by the omission of information \cite{spinde2020b}, our focus is on bias by word choice, in which different words refer to the same concept \cite{recasens2013a}. We will first summarize available media bias data sets and then present automated methods to identify bias as well as MTL. 

The concept of media bias is covered by many data sets \cite{lim2018b, baumer2015a, chen_analyzing_2020, fan2019a,spinde2021mbic}. However, they all exhibit specific deficiencies, such as (1) a low number of topics \cite{lim2018b, lim2020a}, (2) no annotations on the word level \cite{lim2018b}, (3) low inter-annotator agreement \cite{spinde2021mbic, lim2020annotating, baumer2015a, lim2018b}, (4) no background check for its participants (except \cite{spinde2021mbic}), and (5) only article-level annotations \cite{chen_analyzing_2020}. Also, some related papers focus on framing rather than on bias \cite{baumer2015a, fan2019a}, or on Wikipedia instead of news\cite{hube2019neural}, and results are only partially transferable. To the best of our knowledge, the most extensive and precise data set was presented recently \cite{SpindeBABE}. The data set consists of 3700 sentences annotated by expert raters on sentence-level with an inter-annotator agreement of 0.40 measured by Krippendorff's $\alpha$ \cite{krippendorff2011computing}, which is higher than for all other available data sets.

Several studies tackle the challenge of identifying media bias automatically \cite{recasens2013a,Spinde2021,hube2019neural, hube2018detecting, chen_analyzing_2020}. Most of them use hand-crafted features to detect bias \cite{hube2018detecting, Spinde2021}. For example, \cite{Spinde2021} identify and evaluate a wide range of linguistic, lexical, and syntactic features serving as potential bias indicators. The existing work on neural models is based on the data sets mentioned above, which exhibit the described weaknesses \cite{chen_analyzing_2020, hube2019neural}. Most media bias models focus on sentence-level bias \cite{recasens2013a, hube2019neural, hube2018detecting, chen_analyzing_2020, fan2019a}. Therefore, we follow the standard practice and construct a sentence-level classifier. 

MTL approaches have shown to be helpful when high-quality data sets in the domain are scarce, but text corpora covering general related concepts are available \cite{Ernie, liu2019multitask, wang2018glue, sun2019fine, huo2020utilizing}. For example, \cite{huo2020utilizing} report that MTL applied on BERT yields an accuracy increase of 1.03\% compared to the baseline BERT in a subjectivity detection task. MTL might be a suitable training paradigm for media bias identification systems since sufficiently sized bias corpora with qualitative hand-crafted annotations do not exist. Therefore, we propose the first neural MTL media bias classifier composed of inter-domain and cross-domain data sets. 


\section{Methodology}\label{sec:methodology}
We explore how fine-tuning a language model via MTL can improve the performance in detecting media bias on the sentence level. Computational costs are an important consideration for us since we train multiple large-scale MTL models. For this reason, we employ a distilled modification of BERT \cite{devlin2018bert}, called DistilBERT \cite{sanh2020distilbert}, which achieves a 40\% reduction in size while simultaneously accelerating the training process by 60\% and retaining 97\% of language understanding capabilities on NLP benchmark tasks \cite{wang2018glue}. DistilBERT represents an appropriate architecture, keeping resource consumption and performance balanced. The incorporation of larger models trained via MTL is left to future research.

Our MTL technique is based on \textit{hard parameter sharing} in which all hidden model layers are shared between auxiliary training tasks \cite{ruder2017overview}. Task-specific layers are added on top of the last hidden state, accounting for the label structure of auxiliary data sets. The MTL paradigm we propose is architecture-independent and can be adjusted to future NLP architectures.

For our training procedure, we distinguish between models trained on in-domain and cross-domain data sets. For in-domain data sets, the creation process included concepts related to media bias, such as subjectivity \cite{Subj}. Conversely, cross-domain data sets include data points that are not directly annotated for or related to media bias, but are retrieved from tasks that bear some connection to it. The auxiliary data sets we use comprise a diverse set of NLP tasks requiring two different losses for the learning process – the Cross-Entropy (CE) loss \cite{de2005tutorial} and the Mean Squared Error (MSE) loss \cite{ref1}.
The origin and number of the data used for the training of our models, as well as their respective original tasks and used loss functions, are shown in \Cref{table:datasets}. We use in-domain (ID) and cross-domain (CD) data sets used in other MTL studies within the language processing domain \cite{sun2019fine,liu2019multitask,huo2020utilizing}.

\begin{table*}[h]
\centering

\resizebox{0.999\textwidth}{!}{
\begin{tabular}{lccccl}
\hline 
\textbf{Data set} & \textbf{Domain} & \textbf{\textit{n}} & \textbf{Task} & \textbf{Loss} & \textbf{Description}\\
\hline & \\[-1.5ex]
\parbox{4.6cm}{Reddit data set (Reddit) \cite{cabot}} & ID & 6861 & Single Sentence Regression & MSE & \parbox{6.5cm}{Reddit comments labeled on a continuous scale ranging from 0 (supportive) to 1 (discriminatory)} \\[13pt]
\parbox{4.6cm}{Subjectivity data set (Subj) \cite{Subj}} & ID & 10000 & Single Sentence Classification & CE & \parbox{6.5cm}{ Movie reviews labeled as \textit{objective} or \textit{opinionated}}\\[13pt]
\parbox{4.6cm}{IMDb \cite{maas-etal-2011-IMDB}} & ID & 50000 & Single Sentence Classification & CE & \parbox{5cm}{Movie reviews containing positive and negative sentiment labels}\\[13pt]
\parbox{4.6cm}{Wikipedia data set (Wiki)\textsuperscript{1} \cite{Wiki}} & ID & 180000 & Single Sentence Classification & CE & \parbox{6.5cm}{Neutral and biased sentence pairs from articles going against Wikipedia's NPOV rule}\\[13pt]
\hdashline
\parbox{1.2cm}{} & {} &  \parbox{2.2cm}{}\\[5pt]
\parbox{4.6cm}{Semantic Textual Similarity Benchmark
(STS-B) data set \cite{cer2017semeval}} & CD & 10943 & Pairwise Sentence Similarity & MSE &  \parbox{6.5cm}{Multilingual and cross-lingual sentence pairs labeled in terms of similarity}\\[13pt]
\parbox{4.6cm}{Stanford Natural Language Inference (SNLI) corpus \textsuperscript{1} \cite{BowmanAPM15_SNLI}} & CD & 570000 & Pairwise Sentence Classification & CE & \parbox{6.5cm}{Sentence pairs labeled for linguistic relations within the labels \textit{entailment}, \textit{neutral}, or \textit{contradiction}}\\[13pt]
\hline
\end{tabular}
}
\begin{tablenotes}
\small
\item \textsuperscript{1}We only use 50000 text instances from these corpora in our MTL approach to keep the size of training sets balanced.
\end{tablenotes}
\caption{Auxiliary data sets incorporated in the MTL models (\textit{n} = number of instances)} 
\label{table:datasets}

\end{table*}

\begin{figure}
    \centering
    \includegraphics[width = 0.55\columnwidth]{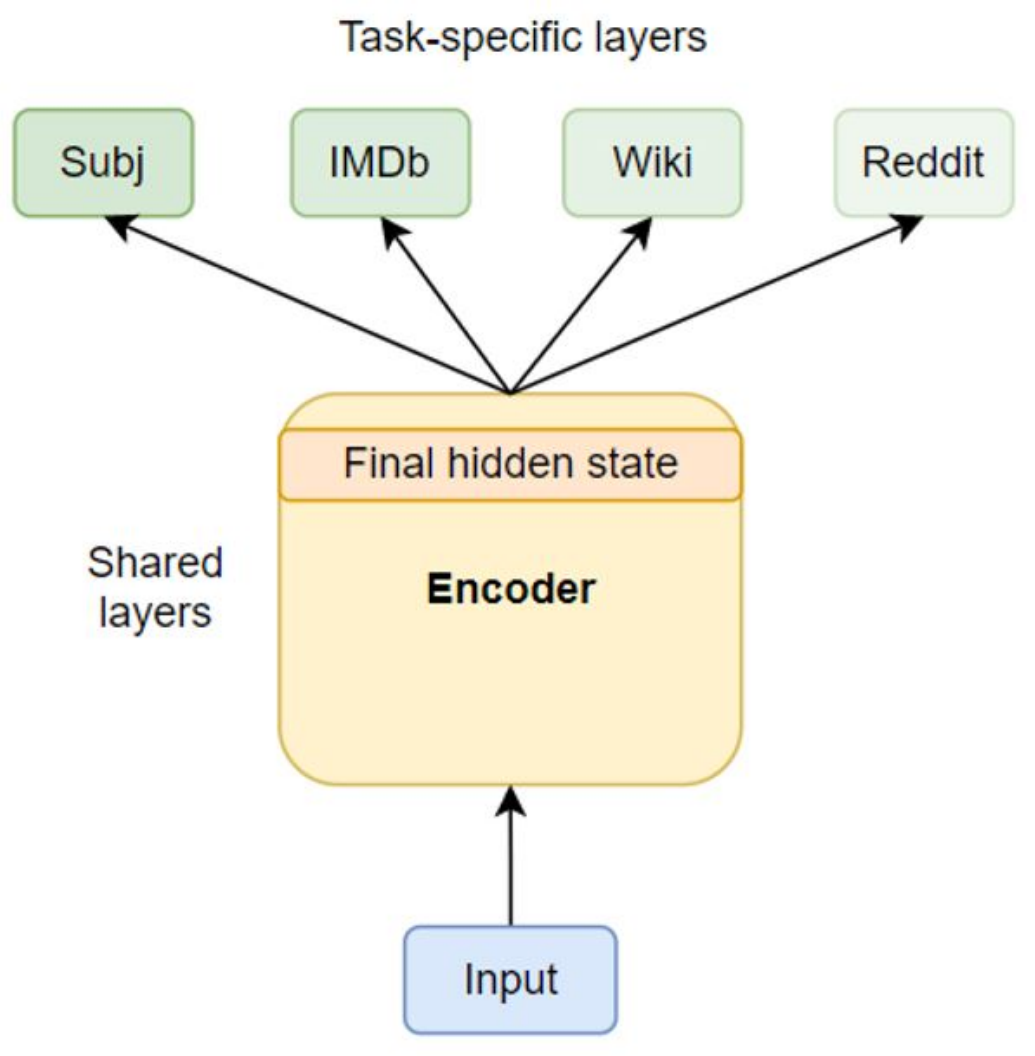}
    \caption{Outline of in-domain  MTL model consisting of a shared encoder block and task-specific layers. \textit{Note}: We implement multiple MTL models based on different combinations of the presented data sets.}
    \label{fig:MTL_model}
\end{figure}

\cref{fig:MTL_model} outlines our in-domain MTL model consisting of DistilBERT's encoder, whose parameters are shared across tasks, and the added task-specific layers \footnote{The cross-domain model is not shown due to lack of space but is published at the repository mentioned in Section \cref{sec:introduction}.}. The represented model is based on the maximum number of possible data sets within the approach. In our experiments on MTL, we try various combinations, including at least three in-domain and five cross-domain data sets, respectively.

For preprocessing and MTL training, we took the same approach as \cite{liu2019multitask}.
Initially, pre-trained parameters are loaded from \textit{huggingface} \footnote{\url{https://huggingface.co/transformers/model_doc/distilbert.html}}. We split up data for a fixed-size subset of tasks into batches, and batches are merged and shuffled to guarantee the model does not train on too many subsequent batches of a single task. The preprocessing step is repeated every epoch. Batches are then passed on by the data loader one by one to the model, which outputs task-specific predictions and the respective loss. Finally, the loss is backpropagated, and parameters are updated. 


\section{Experiments}\label{sec:experiments}
To investigate the benefit of MTL to identify media bias on a fine-grained linguistic level, we train ten models using MTL, which we compare to five baseline models. As a consequence of a lack of robust guidelines for selection criteria for auxiliary corpora, we choose a variety of auxiliary tasks to fine-tune the DistilBERT model via MTL that have previously been used successfully in MTL studies \cite{sun2019fine,huo2020utilizing,liu2019multitask}. Each of our MTL models is trained using a different combination of a sample of six popular data sets, where IMDb \cite{maas-etal-2011-IMDB}, Subj \cite{Subj}, Wiki \cite{Wiki}, and Reddit \cite{cabot} are considered in-domain data sets, and STS-B \cite{cer2017semeval}, and SNLI \cite{BowmanAPM15_SNLI} comprise examples of cross-domain data sets\footnote{A detailed description of the data sets is published at the repository mentioned in Section \cref{sec:introduction}.}.

The in-domain models are based on bias-related data sets \footnote{IMDb, Subj, Wiki, Reddit}. 
Combining the in-domain corpora yields five different models (\cref{fig:results}, M1 - M5): four use triple combinations, and one model relies on all in-domain data sets. The cross-domain models extend the pool of experiments by adding the STS-B and SNLI data sets to each of the five in-domain  models (\cref{fig:results}, M6 - M10). The approaches are oriented on the MTL fine-tuning approaches applied in \cite{sun2019fine}. In their experiments based on BERT, the authors apply MTL on domain-related and domain-unrelated data yielding a performance boost for sentiment classification.

All experiments are performed on a \textit{Google Colab} \textit{NVIDIA Tesla K80} \footnote{\url{https://colab.research.google.com/notebooks/intro.ipynb}}. We choose the \textit{AdamW optimizer} \cite{KingmaB14} and a batch size of 32 . All downstream task layers are based on a hidden state dimensionality of 768. All performance metrics are calculated based on 5-fold cross-validation \cite{Browne_2000}. Thus, we divide the final bias data set containing 1700 instances into five different train and tests \footnote{We use a subset of BABE \cite{SpindeBABE}, introduced in \cref{sec:related_work}, to evaluate the MTL models.}. The models are then iteratively trained on all five training sets and evaluated on the respective held-out test set. Finally, the performance metrics on the test sets are averaged, yielding the cross-validated model performance. Each respective model is trained over four epochs with an early stopping criterion based on validation CE loss. In many cases, the model stops learning after two epochs. The MTL fine-tuning is based on a learning rate of $5\cdot10^{-5}$.

As far as we know, there are no related works applying MTL in the media bias domain. Therefore, we compare the performances of our MTL approaches to a set of baseline models (\cref{fig:results}; B1-B5). We report the performance scores achieved from pre-trained DistilBERT provided by \textit{huggingface} (B1). Furthermore, we train four DistilBERT models on each of the in-domain data sets (B2-B5). Thus, we can observe whether the assumed performance boost of our MTL models results from MTL rather than domain-relatedness of the training data. 

We expect that fine-tuning via MTL leads to an improvement of DistilBERT's bias identification power. Mainly, we want to analyze whether the MTL technique yields a substantial performance boost compared to simple Transfer Learning (TL) approaches training the model on only a single data set. Therefore, we run several experiments. 


\section{Results and Discussion}
We show the performance indicators of our model on our expert-labeled media bias data set in \cref{fig:results}, according to $F_1$, precision, recall, and loss. Since the highest macro $F_1$ score does not necessarily match with the lowest loss, we elaborate on the results from the perspective of both metrics.

\begin{table*}[h]
\centering
\resizebox{1\textwidth}{!}{
\begin{threeparttable}
\begin{tabular}{rlllllllcccccc}
& \textbf{Model} & \multicolumn{6}{l}{\textbf{Data sets}} & \textbf{macro $F_1$} & \textbf{micro $F_1$} & \textbf{binary $F_1$} &	\textbf{Precision} & \textbf{Recall} & \textbf{CE Loss} \\
 \cmidrule{2-14}
& B1 & \multicolumn{6}{c}{\textit{huggingface} DistilBERT} & 0.746 &	0.750 &	0.711 &	\textbf{0.805} & 0.640 &	0.513 \\
\cmidrule{2-14}
\parbox[t]{2mm}{\multirow{4}{*}{\rotatebox[origin=c]{90}{\textbf{TL}}}}
& B2 & \cmark & & & & & & 0.744 & 0.744 & 0.730 & 0.744 & 0.716 & 0.545\\
& B3 & & \cmark & & & & & 0.761 & 0.762 & 0.746 & 0.770 & 0.725 & 0.491\\
& B4 & & & \cmark & & & & 0.743 & 0.746 & 0.709 & 0.790 & 0.646 & 0.497\\
& B5 & & & & \cmark & & & \textbf{0.782} & \textbf{0.782} & \textbf{0.7695} & 0.785 & 0.754 & 0.466\\
\cmidrule{2-14}
\parbox[t]{2mm}{\multirow{5}{*}{\rotatebox[origin=c]{90}{\textbf{ID MTL}}}}
& M1 & \cmark & \cmark & \cmark & & & & 0.768 & 0.768 & 0.753 & 0.778 & 0.731 & 0.482\\
& M2 & \cmark & \cmark & & \cmark & & & 0.760 & 0.760 & 0.746 & 0.766 & 0.729 & 0.495\\
& M3 & \cmark & & \cmark & \cmark & & & 0.773 & 0.774 & 0.762 & 0.777 & 0.755 & 0.482\\
& M4 & & \cmark & \cmark & \cmark & & & 0.776 & 0.777 & 0.759 & 0.794 & 0.727 & \textbf{0.464}\\
& M5 & \cmark & \cmark & \cmark & \cmark & & & 0.772 & 0.771 & 0.757 & 0.778 & 0.737 & 0.473\\
\cmidrule{2-14}
\parbox[t]{2mm}{\multirow{5}{*}{\rotatebox[origin=c]{90}{\textbf{CD MTL}}}}
& M6 & \cmark & \cmark & \cmark & & \cmark & \cmark & 0.766 & 0.766 & 0.758 & 0.756 & 0.763 & 0.492 \\
& M7 & \cmark & \cmark & & \cmark & \cmark & \cmark & 0.765 & 0.765 & 0.751 & 0.770 & 0.735 & 0.474\\
& M8 & \cmark & & \cmark & \cmark & \cmark & \cmark & 0.771 & 0.771 & 0.762 & 0.765 & 0.761 & 0.491\\
& M9 & & \cmark & \cmark & \cmark & \cmark & \cmark & 0.749 & 0.750 & 0.759 & 0.714 & \textbf{0.812} & 0.499\\
& M10 & \cmark & \cmark & \cmark & \cmark & \cmark & \cmark & 0.769 & 0.770 & 0.751 & 0.789 & 0.720 & 0.480\\
\cmidrule{2-14}
& & Subj & IMDb & Reddit & Wiki & STS & SNLI & & & & & &\\
\end{tabular}
\caption{Results for all baseline models, i.e., the \textit{huggingface} model or models obtained by TL, as well as the models trained using MTL considering only in-domain data sets or also incorporating cross-domain data. For each metric we have denoted the best performance in bold.}\label{fig:results}
\end{threeparttable}}
\end{table*}

Among all MTL-trained models the highest $F_1$ score is achieved from the in-domain M4 model with 0.776. The best cross-domain model regarding macro $F_1$ is reached by M8 with 0.771. Compared to DistilBERT, M4 achieves a 3\% increase in macro $F_1$, while B5 achieves the highest macro $F_1$ for TL-based models at 0.782, which is not surpassed by any MTL approach. Although all MTL models outperform DistilBERT, the highest macro $F_1$ score of all MTL models is 0.6\% lower than that of B5. Overall, MTL improves the B1 baseline macro $F_1$ score in a range from 0.3\% (M9) to 3\% (M4). 
When considering the models from a loss-based perspective, the performance ranks change slightly: M4 remains as the best in-domain MTL model, but M7 (the second to last in terms of macro $F_1$ performance) reaches the lowest loss within the cross-domain approaches. Compared to DistilBERT, M4 shows a decrease in loss of 4.9\%. B5 prevails as the best TL model with a CE loss of 0.466. In contrast to the macro $F_1$-based perspective, however, M4 achieves the lowest overall loss, outperforming B5 by 0.2\%.

In general, our MTL approaches surpass the baseline methods. However, the best overall model based on macro $F_1$ was a TL model trained on a data set containing revised Wikipedia excerpts (B5). Based on CE loss, only one MTL model slightly outperform this TL model. Thus, we cannot state whether Transformer-Based MTL improves media bias detection on the sentence level. 
We assume that the strong performance of B5 results from the relatedness of the underlying data set to our bias corpus. The Wikipedia data set contains biased and neutral sentences extracted from revised Wikipedia passages. Hence, the data set is similar to our bias corpus\footnote{Let us point out that \textbf{none} of the instances from the Wikipedia data set are contained in our target media bias data set.}. The only difference to our fine-tuning data set is the source from which the data is extracted. Pre-training a Transformer-based model on a highly bias-related corpus seems to hinder MTL’s relative performance improvement. Furthermore, we assume that our selection of auxiliary data sets might not have been sufficiently comprehensive. In our MTL approaches, updating DistilBERT’s parameters only required the computation and back-propagation of binary CE loss and MSE loss. \cite{ruder2017overview} argues that well-performing MTL approaches must be trained on NLP tasks, including multiple loss functions. 

Existing MTL studies \cite{liu2019multitask,huo2020utilizing,sun2019fine} do not report diverse TL baseline models. The MTL approaches are primarily compared to a pre-trained baseline model provided by model libraries. Future research should incorporate a comprehensive set of baseline models allowing for a more robust analysis. Comparing our best MTL model to DistilBERT, the effect of MTL is similar as in \cite{huo2020utilizing}. 

Considering our MTL-based media bias research, future work should include more comprehensive sets of bias-related auxiliary data sets with multiple loss functions. Possible tasks could, for example, comprise the detection of bias-inducing linguistic features such as negative sentiment \cite{Spinde2021}. In this way, deep learning techniques could benefit from other types of tasks, such as classifying linguistic features.
Moreover, future MTL approaches could benefit from larger transformer models (e.g., XLNet~\cite{YangDYC19}, ELECTRA~\cite{ClarkLLM20}). Our approach based on DistilBERT is the first step towards balancing cloud-computing costs and performance. We note that a follow-up experiment about an improved model and a larger exploratory data analysis are already in progress and will be published in the future. 


\section{Final Considerations}
This work proposes a Transformer-based MTL approach to identify media bias by word choice in news articles. The motivation for selecting the training technique results from our observation that the size of available media bias data sets is not compatible with the requirements of state-of-the-art neural language models. We train ten MTL models based on different combinations of six auxiliary data sets and compare them to five baseline models. Our results show that the best performing MTL model partly surpasses the baseline models in terms of macro $F_1$ loss and CE loss. Yet, we can not ascertain a significant superiority of the MTL approach in classifying media bias instances. The main limitation of our work is the restricted inclusion of auxiliary tasks. In future work, we plan to incorporate more tasks based on bias-inducing linguistic features. We have to emphasize that any successful MTL implementation in the context of media bias identification could decrease financial burdens emerging from the collection of hand-crafted training data. Yet, at the same time, cloud computing requires substantial financial resources. Costs of using larger models should therefore be properly evaluated.
We believe the MTL approach to be promising in the area and aim to continue the research on MTL in connection with media bias identification in the future.
%
%
%
 \bibliographystyle{splncs04}
 \bibliography{mybibliography}

\end{document}